\pdfoutput=1
\documentclass[11pt]{article}

\usepackage{EMNLP2023}

\usepackage{times}
\usepackage{latexsym}

\usepackage[T1]{fontenc}
\usepackage[utf8]{inputenc}

\usepackage{microtype}

\usepackage{inconsolata}

\newcounter{notecounter}
\newcommand{\enotesoff}{\long\gdef\enote##1##2{}}
\newcommand{\enoteson}{\long\gdef\enote##1##2{{
\stepcounter{notecounter}
{\large\bf \hspace{1cm}\arabic{notecounter} $<<<$ ##1: ##2 $>>>$\hspace{1cm}}}}}
\enoteson
\enotesoff % uncomment this to turn off the indexed notes

\def\figref#1{Figure~\ref{fig:#1}}
\def\figlabel#1{\label{fig:#1}\label{p:#1}}

\def\tabref#1{Table~\ref{tab:#1}}
\def\tablabel#1{\label{tab:#1}\label{p:#1}}

\def\secref#1{\S\ref{sec:#1}}
\def\seclabel#1{\label{sec:#1}}
\def\eqref#1{Eq.~\ref{eqn:#1}}

\def\eqlabel#1{\label{eqn:#1}}

\usepackage{xspace}
\usepackage{booktabs}
\usepackage{subcaption}
\usepackage{graphicx}
\usepackage{amssymb}
\usepackage{amsmath}
\usepackage{relsize}
\newcommand{\ra}[1]{\renewcommand{\arraystretch}{#1}}
\usepackage{booktabs}
\usepackage{multirow}
\usepackage{rotating}

\usepackage{mathabx}
\definecolor{cmzhao}{rgb}{0.1, 0.8, 0.1}
\definecolor{c_yuhta}{rgb}{0.831,0.184,0.494}

\definecolor{ballblue}{rgb}{0.13, 0.67, 0.8}

\definecolor{cadetblue}{rgb}{0.37, 0.62, 0.63}

\title{On the Language Encoder of Contrastive Cross-modal Models}

\author{
    Mengjie Zhao\textsuperscript{†}\xspace\xspace
    Junya Ono\textsuperscript{†}\xspace\xspace
    Zhi Zhong\textsuperscript{†}\xspace\xspace
    Chieh-Hsin Lai\textsuperscript{‡}\xspace\xspace
    Yuhta Takida\textsuperscript{‡}\xspace\xspace
    Naoki Murata\textsuperscript{‡}\\
    \textbf{Wei-Hsiang Liao}\textsuperscript{‡}\xspace\xspace
    \textbf{Takashi Shibuya}\textsuperscript{‡}\xspace\xspace
    \textbf{Hiromi Wakaki}\textsuperscript{†}\xspace\xspace
    \textbf{Yuki Mitsufuji}\textsuperscript{‡†}\\
    \textsuperscript{†}Sony Group Corporation \hspace{.5cm}
    \textsuperscript{‡}Sony AI\\
}

\begin{document}
\maketitle

\begin{abstract}

Contrastive cross-modal models such as CLIP and CLAP
aid various vision-language (VL) and audio-language (AL) tasks.
However, there has been
limited investigation of and improvement in
their language encoder, which
is the central component of encoding natural language
descriptions of image/audio into vector representations.
We extensively evaluate how unsupervised and supervised sentence
embedding training affect language encoder quality and
cross-modal task performance.
In VL pretraining, we found that sentence embedding
training  language encoder quality
and aids in cross-modal tasks,
improving contrastive VL models such as CyCLIP.
In contrast, AL pretraining benefits less from
sentence embedding training, which may result from the
limited amount of pretraining data.
We analyze the representation spaces to understand
the strengths of sentence embedding training, and
find that it improves text-space uniformity, at the
cost of decreased cross-modal
alignment.
%\footnote{
%Our code is enclosed in this submission and
%will be open.
%}.
\end{abstract}

\section{Introduction}
Significant progress have been made
in pretraining large-scale cross-modal
models, such as CLIP \citep{radford2021learning}
and ALIGN \citep{jia2021scaling}, for
various vision-language (\textbf{VL}) applications such as
retrieval and zero-shot image classification.
These models are often pretrained with large amounts of data,
e.g., OpenAI leverages  $\approx$400M
caption-image pairs to train CLIP.
The amount of data has been scaled up to
5B with
LAION-5B \citep{schuhmann2022laionb,cherti2022reproducible}.
Such a large amount of multimodal pretraining data
contains text captions at the same scale
as the pretraining corpora of
large language models (\textbf{LLMs})
such as BERT, which is pretrained
on 3.3B words \citep{devlin-etal-2019-bert}.

The success of VL pretraining encourages
research on contrastive learning models for
other modalities like audio.
Pretrained audio-language (\textbf{AL}) models
such as AudioCLIP \citep{audioclip} and
CLAP \citep{laionclap2023,msclap} show promising
results on AL retrieval and zero-shot audio
classification tasks.

It is clear that the language encoder
in cross-modal contrastive models
plays a central role
when scaling-up pretraining of
a specific modality and/or the amount of modalities.
Therefore, analyzing and improving
the language encoder
become increasingly crucial.
CLIP's language encoder (\textbf{CLIP LM}) --
a decoder-only language model similar to
GPT-2 \citep{radford2019language} --
has been investigated.
\citet{yan2022clip}
showed that the CLIP LM outperforms
BERT \citep{devlin-etal-2019-bert} in
clustering entities with prompting.
\citet{wolfe-caliskan-2022-contrastive}
probed the CLIP LM,
showing that word representations from
it are less anisotropic \citep{ethayarajh-2019-contextual}, i.e.,
more uniformly distributed with respect to direction,
than the similar-sized GPT-2.
Complementary to research on
CLIP LM, we focus on pretraining.
CLIP-like models
are often pretrained
with cross-modal contrastive learning.
\emph{
We measure -- during pretraining -- the
effectiveness of systematically
modeling the image captions with
sentence embedding training
} \citep{reimers-gurevych-2019-sentence,gao-etal-2021-simcse},
which is a natural fit to
the captions.

We pretrain CLIP and one
of its new variants CyCLIP \citep{goel2022cyclip}
with sentence embedding training,
as well as the conventional cross-modal contrastive learning.
In addition to CLIP's
cross-modal contrastive learning objective,
CyCLIP explicitly optimizes for geometry
consistency between the
text and image representation spaces,
making it a suitable model for
validating the effectiveness of NLP methods.
We evaluate pretrained models
on an array of tasks involving
one or two modalities such as SentEval,
zero-shot VL retrieval, and image classification.
We find that
unsupervised sentence embedding training
improves the language encoder quality and VL tasks.
Supervised sentence embedding training
improves language encoder quality, but
the benefit does not necessarily
transfer to VL tasks.
We analyze the learned representation
spaces and find that
sentence embedding training
improves text-space
uniformity \citep{wang2020understanding}
and reduces
anisotropy \citep{ethayarajh-2019-contextual,
wolfe-caliskan-2022-contrastive}.

We also investigate
AL contrastive
models such as CLAP \citep{laionclap2023}.
In contrast to VL pretraining,
AL pretraining suffers from data scarcity
and often leverages pretrained LLMs
and audio encoders. We determine the
effectiveness of sentence embedding training
in both scenarios: continued pretraining with
LLMs and audio encoders, and
pretraining from scratch.
We find that
the benefits of sentence embedding
training are less noticeable and noisy.
To the best of our knowledge, this is
the first study on investigating and trying to improve
the language encoder of AL contrastive learning,
and we expect our results will encourage more
research in this direction.

In summary, our \textbf{contributions} are as follows:
(i) We extensively evaluate how
unsupervised and supervised sentence
embedding trainings affect
VL and AL
contrastive pretraining. Experimental
results indicate improved VL performance,
however, the results on AL tasks are noisy
and the improvements are less noticeable.
(ii) We show that unsupervised
sentence embedding training
improves the language encoder of CyCLIP \citep{goel2022cyclip},
hence improves performance of cross-modal tasks.
(iii) We conduct a comprehensive analysis
on the alignment and uniformity of
learned representation spaces
following \citet{wang2020understanding}, and
show that sentence embedding training
improves uniformity of the text representation
space, but at the cost of
decreased cross-modal alignment.

\section{Related work}
\textbf{CLIP LM}.
Research has focused on the language encoder of
OpenAI CLIP \citep{radford2021learning}.
The model consists of a language encoder (CLIP LM) and
an image encoder that are jointly trained on Web-scale
caption-image pairs.
\citet{yan2022clip} stressed the importance of
CLIP LM, showing that it outperforms
BERT \citep{devlin-etal-2019-bert} in tasks, such as
entity clustering, through prompting.
\citet{bielawski-etal-2022-clip} showed that
CLIP LM outperforms BERT
in ``human-centric'' tasks such as genre
classification on books or movies.
\citet{santurkar2023is} highlighted the
importance of text captions for representation
learning of CLIP by comparing it
with SimCLR \citep{chen2020simple} in which
no language supervisions is present.
Training signals from language are shown to be
detrimental, worthing any number of images in a
sufficiently large dataset.
Our work follows this direction, with a focus on
determining how supervised or unsupervised
sentence embedding trainings affect
CLIP LM and VL contrastive learning.

\citet{goel2022cyclip} introduced CyCLIP,
incorporating extra training objectives
than cross-modal contrastive learning
such that the geometry consistency between
the text and image spaces is improved.
One of CyCLIP's training objectives is computing
similarities between captions;
this motivates us to determine how
systematically modeling the captions through
supervised or unsupervised sentence embedding training
affects CyCLIP/CLIP.

\textbf{Contrastive audio-language pretraining}
models have also been proposed. \citet{audioclip} extended
CLIP to audio tasks by adding an extra module and
continued training on audio datasets. Similar distillation
methods such as Wav2CLIP \citep{wavclip}, have also been proposed.
\citet{msclap} and \citet{laionclap2023} independently
proposed CLAP, in which a language encoder and
an audio encoder are jointly trained on AL
datasets, which resembles CLIP.
We focused on the language encoder
in pretraining AL models, and demonstrated
the impact of
sentence embedding training.
To the best of our knowledge, this is the first step
in this direction.

\textbf{Sentence embedding} is an
extensively investigated NLP topic.
Methods ranging from bag-of-word averaging
non-contextualized embeddings \citep{mikolov2013distributed,
pennington2014glove}
to training LSTMs \citep{hochreiter1997long}, e.g.,
SkipThought \citep{kiros2015skip} and
InferSent \citep{infersent},
have been proposed to effectively compose individual tokens
to meaningful sentence representations.
Methods that leverage
post-hoc transforming \citep{li-etal-2020-sentence,su2021whitening}
or finetuning the pretrained BERT
in supervised \citep{reimers-gurevych-2019-sentence}
or unsupervised scenarios \citep{gao-etal-2021-simcse}
are also introduced.
\citet{zhang-etal-2022-mcse} show that
grounding sentence embedding learning
to images improves semantic textual similarity tasks.
We present a
focused investigation on
the LM in CLIP/CyCLIP.
We pretrained from scratch
an LM and ResNet-50 \citep{he2016deep}
with cross-modal contrastive learning,
as well as
unsupervised or supervised
sentence embedding training with image captions.
Verifying the effectiveness of
sentence embeddings -- a critical component for retrieval
and clustering \citep{reimers-gurevych-2019-sentence,
gao-etal-2021-simcse,wang-etal-2021-tsdae-using,
thakur2021beir,GeigleRetri} --
is of great importance
because retrieval has been one of the
main applications of CLIP-like models.

\section{Method}
\seclabel{sec:method}
Cross-modal contrastive learning plays a key role in
training models such as CLIP/CLAP.
We take image modality as an example for introducing this  method.
Consider a caption-image dataset $\{(I_i, T_i)\}_{i=1}^{N}$
that includes $N$ caption-image pairs, and denote $I^e$ and $T^e$ as the
output representations from an image and langauge encoder, respectively.

The cross-modal contrastive loss \citep{radford2021learning}
is defined as
\vspace{-.3cm}
\begin{align*}
\eqlabel{cliploss}
 \mathcal{L_{\textnormal{contra.}}(\tau)} = -\sum_{j = 1}^{N}\log &\frac{\exp\left({\langle}I^{e}_{j}, T^{e}_{j}{\rangle}/\tau\right)}{\sum_{k}{\exp\left({\langle}I^{e}_{j}, T^{e}_{k}{\rangle}/\tau\right)}} \nonumber \\
                                &\hspace{-2cm}\quad -\sum_{k = 1}^{N}\log\frac{\exp\left({\langle}I^{e}_{k}, T^{e}_{k}{\rangle}/\tau\right)}{\sum_{j}{\exp\left({\langle}I^{e}_{j}, T^{e}_{k}{\rangle}/\tau\right)}},
\end{align*}
\vspace{-.3cm}

\noindent where $\tau$ is a trainable temperature parameter
initialized to 0.07
and $\langle\cdot,\cdot\rangle$ computes cosine similarity.

CLIP's training objective solely stresses the alignment
between the two modalities.
CyCLIP \citep{goel2022cyclip} has improved CLIP \citep{radford2021learning}
by additionally optimizing for improved representation space
geometry, such that the image and text spaces are more consistent
with each other. Concretely,
CyCLIP explicitly optimizes two additional
objectives for
cross-modal and in-modal consistency
as well as $\mathcal{L_{\textnormal{contra.}}}$:
\begin{align*}
\mathcal{L_{\textnormal{C-cyclic}}} = \sum_{j}\sum_{k}\left({\langle}I^{e}_{j}, T^{e}_{k}{\rangle} - {\langle}I^{e}_{k}, T^{e}_{j}{\rangle}\right)^{2}, \\
\mathcal{L_{\textnormal{I-cyclic}}} = \sum_{j}\sum_{k}\left({\langle}I^{e}_{j}, I^{e}_{k}{\rangle} - {\langle}T^{e}_{k}, T^{e}_{j}{\rangle}\right)^{2}.
\end{align*}

Intuitively, decreasing the cross-modal consistency loss
$\mathcal{L}_{\textnormal{C-cyclic}}$ makes the cross-modal similarity matrix
more symmetric, as shown in \figref{crossmodalconsistency}.
Note that solely optimizing
$\mathcal{L_{\textnormal{contra.}}}$ is expected to
symmetrize the cross-modal similarity matrix because
the similarity of non-diagonal pairs are trained to be zero.
\citet{goel2022cyclip} showed that this scenario does not occur
in practice and explicitly optimizing $L_{\textnormal{C-cyclic}}$ is
beneficial.

\begin{figure}[t]
    \centering
    \includegraphics[width=.7\linewidth]{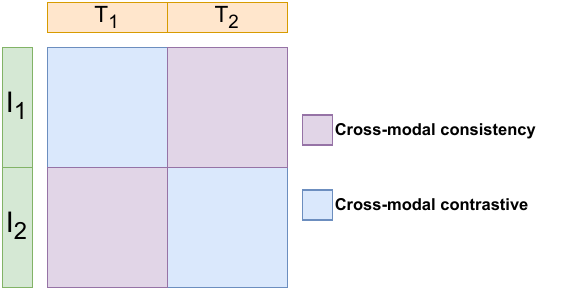}
    \caption{
    Cross-modal similarity matrix. Diagonal elements refer to
    cosine similarity between aligned caption-image pairs while non-diagonal
    elements refer to mismatched caption-image pairs.
    }
    \figlabel{crossmodalconsistency}
\end{figure}

Optimizing $\mathcal{L}_{\textnormal{I-cyclic}}$, however,
reduces the inconsistency between
the overall geometry of the text and image spaces.
Computing $\mathcal{L}_{\textnormal{I-cyclic}}$
requires calculating the similarity
between two text captions,
i.e., ${\langle}T^{e}_{k}, T^{e}_{j}{\rangle}$.
It is thus reasonable to hypothesize that
\emph{accurately computing caption similarities is beneficial for
optimizing} $\mathcal{L}_{\textnormal{I-cyclic}}$. We use
unsupervised and supervised sentence embedding training methods and
test the hypothesis (\secref{sec:exp}).

We use the widely used SimCSE \citep{gao-etal-2021-simcse}
method for learning caption representations.
SimCSE also uses contrastive learning:
a caption is input to a language encoder twice to obtain
two vectors $T^{e}$ and $T^{e}_{+}$.
With dropout enabled, $T^{e}$ and $T^{e}_{+}$
are generally different. These
paired vectors serve as
the positive training pairs
for contrastive learning,
while mismatched captions form negative pairs.
We denote the unsupervised SimCSE loss as
\begin{align*}
 \mathcal{L_{\textnormal{s}}(\tau)} = -\sum_{j}\log &\frac{\exp\left({\langle}T^{e}_{j}, T^{e}_{j,+}{\rangle}/\tau\right)}{\sum_{k}{\exp\left({\langle}T^{e}_{j}, T^{e}_{k,+}{\rangle}/\tau\right)}}, \nonumber
\end{align*}
\noindent where $\tau$ is fixed to 0.05 following \citep{gao-etal-2021-simcse}.

Another direction of sentence embedding training is
supervised training \citep{reimers-gurevych-2019-sentence} on
natural language inference (NLI)
datasets, e.g., SNLI and MNLI
\citep{bowman-etal-2015-large,mnlidata}.
We denote the objective as $\mathcal{L_{\textnormal{n}}}$
and follow \citet{gao-etal-2021-simcse} in
using entailment pairs as $T^{e}$ and $T^{e}_{+}$ and
the contradiction sentence as a hard negative.

\tabref{listoflosses} lists various models we use in our
experiments as well as their training objectives.
During the
experiments, we sum up the objectives but weight
them with different
hyperparameters ($\lambda$) depending on the combinations,
which are shown in \secref{sec:exp}.
We add a suffix ``s'' to the name of models trained with
$\mathcal{L_{\textnormal{s}}}$
and ``n'' to models trained with $\mathcal{L_{\textnormal{n}}}$.

\begin{table}[t]
\centering
\footnotesize
\ra{.9}
\begin{tabular}{@{}lccccc@{}}\toprule
          & $\mathcal{L_{\textnormal{contra.}}}$ & $\mathcal{L_{\textnormal{C-cyclic}}}$& $\mathcal{L_{\textnormal{I-cyclic}}}$ & $\mathcal{L_{\textnormal{s}}}$ & $\mathcal{L_{\textnormal{n}}}$  \\
\cmidrule{1-6}
CLI (A) P      & \checkmark              & -                    & -                   & -             & - \\
CLI (A) Ps     & \checkmark              & -                    & -                   & \checkmark    & - \\
CLI (A) Pn     & \checkmark              & -                    & -                   & -             & \checkmark \\
CyCLI (A) P    & \checkmark              & \checkmark           & \checkmark          & -             & - \\
CyCLI (A) Ps   & \checkmark              & \checkmark           & \checkmark          & \checkmark    & - \\
CyCLI (A) Pn   & \checkmark              & \checkmark           & \checkmark          & -             & (\checkmark) \\
\bottomrule
\end{tabular}
\caption{List of training objectives.
We follow \citet{radford2021learning} for CLIP, \citet{laionclap2023} for CLAP,
and \citet{goel2022cyclip} for CyCLIP.}
\tablabel{listoflosses}
\end{table}
\normalsize

\section{Experiments}
\seclabel{sec:exp}
\subsection{Datasets}
\seclabel{sec:expdataset}
To \emph{pretrain VL models} such as CLIP and CyCLIP,
we follow \citet{bugliarello-etal-2021-multimodal,goel2022cyclip}
and use the Conceptual Captions dataset,
which consists of approximately\footnote{
CC3M images need to be downloaded by users.
Due to broken URLs,
the exact amount of data varies from time to time;
\tabref{datasetstats} shows the exact number of images.}
3M caption-image pairs
(CC3M; \citet{sharma-etal-2018-conceptual}).
CC3M has a reasonable size for pretraining and
contains a broad coverage of Web content, making
it a good option for learning
generic VL representations
\citep{bugliarello-etal-2021-multimodal}.

To \emph{evaluate the
trained VL models},
we follow \citep{radford2021learning} and
conduct evaluations
with zero-shot image-text retrieval on
the Karpathy \citep{karpathy2015deep} test splits
of Flickr30K \citep{flikr} and
MSCOCO \citep{cococaption}.
We skip the evaluation on Flickr30K
when supervised sentence embedding
training is used, i.e.,
when $\mathcal{L_{\textnormal{n}}}$
is considered in training. This is because
Flickr30K captions are
the premises in the SNLI dataset \citep{bowman-etal-2015-large},
overlapping with the supervised sentence embedding training data.
For zero-shot image classification,
we use the standard benchmarks
CIFAR10, CIFAR100 \citep{cifar},
and ImageNet1K \citep{russakovsky2015imagenet}.
Zero-shot image classification
with domain shift, out-of-domain, and adversarial examples
are also considered:
ImageNetV2 \citep{imagenetv2},
ImageNet-Sketch \citep{imagenetsketch},
ImageNet-O, ImageNet-A, and
ImageNet-R \citep{hendrycks2021natural,hendrycks2021many}.

To \emph{pretrain AL models}, e.g.,
CLAP and CyCLAP, we conduct experiments with Clotho \citep{clotho}
consisting of $\approx$6K caption-audio pairs and AudioCaps consisting of
$\approx$50K\footnote{
Similar to CC3M, AudioCaps only provides audio captions while
users need to download  corresponding YouTube videos
and convert their audio
to waveforms; \tabref{datasetstats} shows the exact
amount of waveforms we used.
} caption-audio pairs \citep{kim-etal-2019-audiocaps},
similar to LAION-CLAP \citep{laionclap2023}.  In
contrast to the VL scenario,
AL pretraining is known to
be challenging due to data scarcity \citep{laionclap2023}.
We use Clotho and AudioCaps because they contain
natural language descriptions of the audio instead of
simple tags; thus, there is no need to generate
pseudo-natural language captions from tags using
services such as ChatGPT/GPT3 \citep{mei2023WavCaps}.

To \emph{evaluate
pretrained AL models},
we conduct cross-modal retrieval and
zero-shot audio classification tasks.
For Clotho, we train the models on the training
split and report retrieval results on the validation split.
For AudioCaps, we select the best-performing checkpoint on
the validation split and report test split results.
We conduct zero-shot classification
on the Environmental Sound Classification dataset (ESC50; \citet{ESCdata})
and UrbanSound8K (US8K; \citet{us8k}),
which have been widely
used \citep{laionclap2023,msclap}.
ESC50 contains
short audio clips containing the sound of
different common events such as cats meowing and
dogs barking; the clips are categorized into 50 classes,
and US8K contains audio clips
of urban event sounds such as drilling and street music;
the clips are categorized into ten classes.

\begin{table}[t]
\centering
\scriptsize
\ra{1.1}
\begin{tabular}{@{}llcccr@{}}
\toprule
                                & \multicolumn{1}{l}{Dataset} & Pretraining & Retrieval & ZS          & Size \\ \midrule
                                & CC3M                        & \checkmark   & -         & -          & 2,806,641     \\
                                & MSCOCO                      & -           & \checkmark & -          & 5,000     \\
                                & Flickr30K                   & -           & \checkmark & -          & 1,000   \\
                                & CIFAR10                     & -           & -         & \checkmark  & 10,000     \\
                                & CIFAR100                    & -           & -         & \checkmark  & 10,000     \\
\textbf{VL    }                 & ImageNet1K                  & -           & -         & \checkmark  & 50,000   \\
                                & ImageNetV2                  & -           & -         & \checkmark  & 10,000     \\
                                & ImageNetSketch              & -           & -         & \checkmark  & 5,0889     \\
                                & ImageNet-O                  & -           & -         & \checkmark  & 2,000   \\
                                & ImageNet-A                  & -           & -         & \checkmark  & 7,500     \\
                                & ImageNet-R                  & -           & -         & \checkmark  & 30,000     \\ \midrule
\multirow{4}{*}{\textbf{AL}}    & Clotho                      & \checkmark   & \checkmark & -         & 5,929 \\
                                & AudioCaps                   & \checkmark   & \checkmark & -         & 50,725  \\
                                & ESC50                       & -           & -         & \checkmark  & 400     \\
                                & US8K                        & -           & -         & \checkmark  & 8,732  \\ \bottomrule
\end{tabular}
\caption{Datasets and their amount of examples.
We report amount of images for VL datasets and of
waveform files for AL datasets.
``ZS'': zero-shot classification.
}
\tablabel{datasetstats}
\end{table}

\tabref{datasetstats} lists all the cross-modal datasets and their usage.
We follow \citet{goel2022cyclip} in processing the VL
datasets and \citet{laionclap2023} in processing the AL datasets;
the detailed steps of these processes are shown
in Appendix \secref{appendixsec:dataandhyper}.

\subsection{Experiment settings}
To \emph{pretrain the VL models},
we use the same model architecture
as CyCLIP \citep{goel2022cyclip}, i.e.,
a ResNet-50 as the image encoder
and Transformer \citep{vaswani2017attention} as
the language encoder.
We pretrain the model from scratch and largely
reuse CyCLIP's hyperparameters.
To weight different
training objectives (\tabref{listoflosses})
in CyCLIP, we set $\lambda$\textsubscript{I-cyclic} and
$\lambda$\textsubscript{C-cyclic} to 0.25,
$\lambda$\textsubscript{contra.} is set to 1.0, and we empirically set
$\lambda$\textsubscript{s}
and $\lambda$\textsubscript{n} to 0.1.
We use a batch size of 80, and each
pretraining trial is run for 64 epochs,
taking four days with four A100 GPUs.
We enable dropout in the language encoder and
use dropout rate of 0.1.
Appendix \secref{appendixsec:dataandhyper} lists the
details of the hyperparameters.

The VL pretraining dataset CC3M has a validation split consisting of
$\approx$15K text-image pairs. We select the best-performing
checkpoint on this validation split for downstream task evaluations.

Due to data scarcity, \emph{AL pretraining} often
leverages pretrained language and audio encoders.
We use the same model
architecture as LAION-CLAP \citep{laionclap2023}: pretrained
RoBERTa-base \citep{liu2020roberta} as the language encoder
and pretrained Hierarchical Token-Semantic Audio Transformer (HTSAT;
\citet{chen2022hts}) as the audio encoder. HTSAT has shown to
outperform CNNs in various audio tasks \citep{chen2022hts}.
We use the \texttt{HTSAT-tiny} variant with 31M parameters.
Due to small dataset size, each
pretraining experiment takes less than
one day on a single A100 GPU.
We use the default dropout rate of 0.1 of
RoBERTa-base in our experiments.
In pilot experiments on Clotho, we found that the
default hyperparameters\footnote{
\url{https://github.com/LAION-AI/CLAP/blob/main/experiment_scripts/train-only-clotho.sh}}
lead to
underfitting\footnote{This is reflected by the
increasing validation performance in learning curves
publicly shared by the LAION-CLAP project on \href{https://stability.wandb.io/clap/clap/reports/CLAP-trained-on-Clotho-dataset--VmlldzoyNzY?accessToken=c0erq9hhp7h880jclihd9j9if679s6bylwto33vo14yo5jg40ppe38qeoafoonpz}{WnB}.
}.
We find that longer training and a smaller learning rate
improve performance, as shown in
\tabref{map10val}. \tabref{alhyper} lists our hyperparameters,
which are kept the same
when we pretrain the AL models on Clotho or AudioCaps.

For weighting the training objectives,
due to the small model size and dataset size, we grid search
the optimal
$\lambda$\textsubscript{I-cyclic},
$\lambda$\textsubscript{C-cyclic},
and $\lambda_{\textnormal{s}}$
from [0.1, 0.25, 0.5];
$\lambda$\textsubscript{contra.}
is set to 1.0.
Supervised sentence embedding training
($\lambda$\textsubscript{n}) is not considered due to the small size
of the AL datasets.

Clotho and AudioCaps both contain validation splits; thus, we
select the best-performing checkpoint on the validation splits
then conduct evaluation on downstream tasks.

\section{Results and analyses}
\subsection{Results}
\seclabel{subsec:res}

\begin{table}[t]
\hspace{-1.3cm}
\scriptsize
\centering
\begin{subtable}[b]{0.4\textwidth}
\centering
\ra{1.3}
\begin{tabular}{@{}lrrrp{.15cm}rrrc@{}}\toprule
     & \multicolumn{3}{c}{Text Retrieval} & \phantom{abc}& \multicolumn{3}{c}{Image Retrieval}  \\
\cmidrule{2-4} \cmidrule{6-8}
         & R@1   & R@5   & R@10  && R@1   & R@5   & R@10  \\ \midrule
CLIP     & 15.70 & 37.22 & 49.06 && 12.48 & 31.10 & 42.23 \\
CLIPs    & \textbf{17.78} & \textbf{38.92} & \textbf{50.10} && \textbf{13.46} & \textbf{32.93} & \textbf{44.09} \\
CLIPn    & 15.74 & 35.66 & 47.38 && 13.12 & 31.46 & 42.55 \\\midrule
CyCLIP   & 18.92 & 41.46 & 54.00 && 15.40 & 35.61 & 46.95 \\
CyCLIPs  & \textbf{21.30} & \textbf{44.34} & \textbf{56.54} && \textbf{16.69} & \textbf{37.75} & \textbf{49.24} \\
CyCLIPn  & 16.32 & 36.76 & 48.16 && 14.53 & 34.07 & 45.52 \\
%%\bottomrule
\end{tabular}
\end{subtable}
\vspace{1em}\quad\\
\hspace{-1.3cm}
\begin{subtable}[b]{0.4\textwidth}
\ra{1.2}
\begin{tabular}{@{}lrrrp{.15cm}rrrp{.15cm}rrr@{}}%%\toprule
     & \multicolumn{3}{c}{Text Retrieval} & \phantom{abc}& \multicolumn{3}{c}{Image Retrieval}  \\
\cmidrule{2-4} \cmidrule{6-8}
         & R@1   & R@5   & R@10  && R@1   & R@5   & R@10  \\ \midrule
CLIP     & 31.80 & 62.10 & 72.90 && 25.50 & 52.28 & \textbf{64.34} \\
CLIPs    & \textbf{35.20} & \textbf{63.20} & \textbf{75.30} && \textbf{26.70} & \textbf{52.34} & 64.32\\\midrule
CyCLIP   & 37.30 & 66.10 & 76.40 && 30.22 & 56.70 & 67.40\\
CyCLIPs  & \textbf{40.00} & \textbf{69.30} & \textbf{79.70} && \textbf{31.74} & \textbf{58.02} & \textbf{69.46}\\
\bottomrule
\end{tabular}
\end{subtable}
\caption{
Zero-shot \textbf{VL retrieval results} (\%)
on MSCOCO (top) and Flickr30K (bottom).
Integrating unsupervised sentence embedding
learning (CLIPs and CyCLIPs)
noticeably improves zero-shot retrieval;
supervised embedding training (CLIPn and CyCLIPn)
has neutral to negative impacts.
}
\tablabel{retrievalres}
\end{table}
\normalsize

\begin{table}[t]
\scriptsize
\centering
\begin{tabular}{@{}lcc@{}}
\toprule
mAP@10  & Text Retrieval & Audio Retrieval \\ \midrule
CLAP    & 12.24          & \textbf{21.21}           \\
CLAP \citep{laionclap2023} & \textbf{13.80}          & 20.40 \\ \midrule
CLAP    & \textbf{51.34}          & \textbf{55.60}           \\
CLAP \citep{laionclap2023} & 45.70          & 51.30 \\
\bottomrule
\end{tabular}
\caption{
AL training validation performance (\%).
We report mean average precision at ten (mAP@10) on
validation split of
Clotho (top) and AudioCaps (bottom).
}
\tablabel{map10val}
\end{table}
\normalsize

\tabref{retrievalres} lists the
\textbf{zero-shot VL retrieval}
results on MSCOCO and Flickr30K,
following the settings in \citet{radford2021learning}.
We first utilize the pretrained
image and language encoders in a VL model
to encode images and captions into vectors.
In text retrieval, we then input the image vector
to retrieve the aligned captions and vice-versa for
image retrieval. We report Recall@N for N in [1, 5, 10].

\textbf{Comparing CyCLIP and CLIP variants.}  We see that
CyCLIP clearly outperforms CLIP across the board for both datasets,
highlighting the significant value of incorporating
$\mathcal{L_{\textnormal{C-cyclic}}}$ and
$\mathcal{L_{\textnormal{I-cyclic}}}$ in optimizing for consistent
geometry of the text and image representation spaces
\citep{goel2022cyclip}.

When \textbf{comparing CyCLIPs/CLIPs to CyCLIP/CLIP},
we observe the effectiveness of improving
the language encoder with
unsupervised sentence embedding
training $\mathcal{L_{\textnormal{s}}}$.
CyCLIPs/CLIPs clearly surpass CyCLIP/CLIP in
all configurations except Flickr30K-CLIPs-ImageRetrieval-R@10.
This suggests that, \emph{we improve CyCLIP on zero-shot vision-language retrieval
tasks through learning better representations of the captions}.
We also observe more gains in text retrieval
than in image retrieval. For example, on Flickr30K@1,
CyCLIPs outperforms CyCLIP by 2.70\% (absolute) in text retrieval
and by 1.52\% (absolute) in image retrieval. This observation reflects
the effectiveness of improved caption representations.

In contrast to the unsupervised embedding training scenario
($\mathcal{L_{\textnormal{s}}}$ and CyCLIPs/CLIPs),
supervised sentence embedding training ($\mathcal{L_{\textnormal{n}}}$
and CyCLIPn/CLIPn) results in a neutral to negative impact on the
overall retrieval results.  Investigating the text and image
representation spaces, we find that $\mathcal{L_{\textnormal{n}}}$
extensively enforces a uniform text representation space such that the
alignment between the text and image spaces is negatively affected; we
provide more in-depth analyses in \secref{alignmentuniform}.

\begin{table}[t]
\scriptsize
\begin{subtable}[b]{0.4\textwidth}
\hspace{-.15cm}
\ra{1.2}
\begin{tabular}{@{}lrrrp{.05cm}rrrc@{}}\toprule
     & \multicolumn{3}{c}{Text Retrieval} & \phantom{abc}& \multicolumn{3}{c}{Audio Retrieval}  \\
\cmidrule{2-4} \cmidrule{6-8}
         & R@1   & R@5   & R@10  && R@1   & R@5   & R@10  \\ \midrule
CLAP     & \textbf{13.88} & 34.16 & 48.90 && 11.67 & \textbf{33.80} & \textbf{47.10} \\
CLAPs    & 13.49 & \textbf{35.60} & \textbf{49.00} && \textbf{11.92} & 32.54 & 45.47 \\ \midrule
CyCLAP   & 14.74 & 35.50 & 48.52 && 11.90 & \textbf{34.95} & \textbf{48.61} \\
CyCLAPs  & \textbf{14.93} & \textbf{36.84} & \textbf{51.00} && \textbf{12.08} & 34.09 & 46.76 \\
%%\bottomrule
\end{tabular}
\end{subtable}
\vspace{1em}\quad\\
\begin{subtable}[b]{0.4\textwidth}
\hspace{-.15cm}
\ra{1.3}
\begin{tabular}{@{}lrrrp{.05cm}rrrcrrr@{}}%%\toprule
     & \multicolumn{3}{c}{Text Retrieval} & \phantom{abc}& \multicolumn{3}{c}{Audio Retrieval}  \\
\cmidrule{2-4} \cmidrule{6-8}
         & R@1   & R@5   & R@10  && R@1   & R@5   & R@10  \\ \midrule
CLAP     & \textbf{44.10} & \textbf{76.80} & \textbf{87.67} && \textbf{34.82} & \textbf{70.62} & 82.93 \\
CLAPs    & 42.73 & 75.44 & 87.57 && 34.69 & 69.80 & \textbf{82.99} \\\midrule
CyCLAP   & \textbf{40.65} & 74.19 & \textbf{86.52} && 34.13 & 69.24 & 82.30\\
CyCLAPs  & 39.81 & \textbf{74.40} & 85.79 && \textbf{34.23} & \textbf{70.24} & \textbf{82.74}\\
\bottomrule
\end{tabular}
\end{subtable}
\caption{
\textbf{Text-audio retrieval results} (\%)
on Clotho (top) and AudioCaps (bottom).
Adding unsupervised sentence embedding training
improves performance of CLAP/CyCLAP in general,
but improvements are noisy and less noticeable than in
VL scenario.
}
\tablabel{audioretrievalres}
\end{table}
\normalsize

\textbf{AL retrieval results} on Clotho and AudioCaps are
listed in \tabref{audioretrievalres}. Supervised sentence embedding
training objective $\mathcal{L_{\textnormal{n}}}$ is not considered
because NLI datasets are much larger than AL datasets
(MNLI: 433K; Clotho: 6K; AudioCaps: 53K); subsampling introduces
extra random factors that are difficult to control.

\textbf{On Clotho},
we observe overall improvements when comparing CyCLAP to CLAP.
We see one exception on text-retrieval-R@10 (
i.e., 48.90 for CLAP and 48.52 for CyCLAP);
however, the overall improvement
outweighs performance drops, demonstrating
that explicitly optimizing for the consistency
between the audio and text spaces \citep{goel2022cyclip} is
also promising for improving AL retrieval on Clotho.
When comparing CyCLAPs/CLAPs with CyCLAP/CLAP,
we see clear improvements on text retrieval.
However, this comes with a decreased
performance on audio retrieval results.

\textbf{On AudioCaps}, CyCLAP falls behind CLAP. This is somewhat
surprising because AudioCaps is considered an easier dataset than
Clotho \citep{audioretrievaljournal}.
The HTSAT audio
encoder has already been pretrained with audio
classification tasks on
AudioSet \citep{gemmeke2017audio},
from which AudioCaps is derived.
This may contribute to the noisy results.
For example, LAION-CLAP \citep{laionclap2023} reported
that adding additional 630K AL pairs
largely boosts AL retrieval performance on Clotho,
but hurts on AudioCaps.
We observe similar results when comparing
CyCLAPs/CLAPs with CyCLAP/CLAP.

\textbf{Comparing VL and AL retrieval results} in
\tabref{retrievalres} and \tabref{audioretrievalres}, we observe that
(1) CyCLIP noticeably improves over CLIP than CyCLAP
over CLAP, and CLAP even outperforms CyCLAP on AudioCaps;
and (2) improving the language encoder with sentence embedding training
is more beneficial to VL than AL.
We hypothesize that this is because AL pretraining starts with
pretrained encoders, which have
geometry that is difficult to alter due to the small AL dataset size.
We further conduct AL pretraining from scratch (Appendix
\secref{appendix:scratchaudio}), where the language and audio encoders
are re-initialized.
Sentence embedding training brings more noticeable results,
however, the the overall results are still
noisy due to the particularly small dataset sizes.
We believe resolving the data scarcity issue is a critical step
towards clearer regularities in AL pretraining.
Researchers have started to exploit directions
like using ChatGPT \citep{mei2023WavCaps}.

\begin{table}[t]
\centering
\tiny
\ra{1.2}
\begin{tabular}{@{}lrrr|rrr@{}}\toprule
                & CLIP  &CLIPn& CLIPs  & CyCLIP & CyCLIPn & CyCLIPs \\
\cmidrule{1-7}
CIFAR10         & 28.31 &\textbf{44.06} & 36.80 & 38.67  & 41.16   & \textbf{44.97} \\
CIFAR100        & 13.23 &\textbf{17.93} & 10.72 & 17.44  & 19.82   & \textbf{22.05} \\
ImageNet1K      & 14.94 &15.97 & \textbf{16.01} & 20.99  & 18.13   & \textbf{22.13} \\\midrule
ImageNetV2      & 12.85 &13.41 & \textbf{14.09} & 17.77  & 15.65   & \textbf{18.68} \\
ImageNet-Sk.    & 7.72  &7.75  & \textbf{8.14}  & 11.67  & 9.93    & \textbf{12.85} \\
ImageNet-O      & 20.75 &\textbf{21.95} & 21.30 & 27.05  & 24.45   & \textbf{29.55} \\
ImageNet-A      & 3.59  &3.41  & \textbf{3.95}  & 5.03   & 4.45    & \textbf{5.19} \\
ImageNet-R      & 18.39 &\textbf{18.51} & 18.24 & 24.37  & 23.07   & \textbf{26.72} \\
\bottomrule
\end{tabular}
\caption{
Zero-shot image classification (R@1 in \%) on
standard datasets (top) and
datasets with distribution shift
or adversarial examples (bottom).
}
\tablabel{zeroshotclassification}
\end{table}
\normalsize

For \textbf{zero-shot image classification},
\tabref{zeroshotclassification} lists the Top1 accuracy
on standard image classification datasets (top) and
datasets with distribution shifts or
adversarial examples (bottom).
We follow \citet{radford2021learning}
and use their prompts for zero-shot classification.
For an image to be classified,
we compute the cosine similarity between its
vector and the encoded vector of all 
classes. Each of the classes is reformulated
with various prompts.
E.g., the ImageNet class ``plane''
is reformulated with 80
templates \footnote{OpenAI templates
are public on
\href{https://github.com/openai/CLIP/blob/main/notebooks/Prompt_Engineering_for_ImageNet.ipynb}{GitHub link}.}
such as
``a photo of a '' and ``a blurry photo of a '',
resulting in prompts
``a photo of a plane'' and
``a blurry photo of a plane''
\citep{radford2021learning}.
The vectors of encoded prompts of a class are averaged;
we select the class
with the maximum  cosine similarity with the image vector.

Similar trends as in the retrieval tasks are observed.
CyCLIP variants outperform their CLIP counterparts;
unsupervised sentence embedding training benefits
both CyCLIP/CLIP while supervised
sentence embedding training does not result in
consistent improvement or deterioration.

For \textbf{zero-shot audio classification},
\tabref{audiozeroshotclassification} list the Top1
accuracy on ESC50 and US8K of models pretrained on
Clotho and AudioCaps respectively.
We follow LAION-CLAP \citep{laionclap2023}
and use their template ``This is a sound of ''.
In contrast to the image scenario, only one
template is provided.
CyCLAP generally outperforms CLAP, except
for AudioCaps-US8K.
We observe no improvements of
sentence embedding training when comparing CyCLAPs/CLAPs
with CyCLAP/CLAP.
We leave the extensive ``prompt engineering'' of
designing more prompts for future work\footnote{
Following the VL scenario, we write prompts and conduct
more experiments on zero-shot audio classification.
Appendix \secref{appendixsec:audiozeronewprompt} shows
results and discussions.
We see improved performance,
however, the generalization ability of this prompt engineering
needs to be tested, and we leave it for future work.
}.

\begin{table}[t]
\ra{1.1}
\scriptsize
\centering
\begin{tabular}{@{}llrr|rr@{}}
\toprule
                           &       & CLAP  & CLAPs & CyCLAP & CyCLAPs \\ \midrule
\multirow{2}{*}{Clotho}    & ESC50 & \textbf{72.25} & 71.75 & \textbf{78.00}  & 75.75      \\
                           & US8K  & \textbf{67.76} & 67.42 & \textbf{70.46}  & 65.98      \\ \midrule
\multirow{2}{*}{AudioCaps} & ESC50 & \textbf{74.50} & 72.75 & \textbf{77.50}  & 75.50        \\
                           & US8K  & \textbf{69.78} & 64.14 & \textbf{67.90}  & 66.94     \\ \bottomrule
\end{tabular}
\caption{
Zero-shot audio classification (R@1 in \%)
on ESC50 and US8K of models pretrained
on Clotho or AudioCaps.
}
\tablabel{audiozeroshotclassification}
\end{table}

\subsection{Analyses}
\seclabel{alignmentuniform}
\textbf{Alignment and uniformity of representation spaces}.
In this section, we take a closer look at the learned representation
spaces.  Following \citet{wang2020understanding}, we inspect the
alignment and uniformity on the hypersphere of learned spaces.
Considering a caption-image dataset $\{(I_i, T_i)\}_{i=1}^{N}$,
we can compute the alignment and uniformity scores defined as:
\begin{equation*}
\begin{gathered}
\mathcal{L}_{\text{align}}\triangleq \mathop{\mathbb{E}}_{{\tiny{(I, T)}}~\sim~ p_{\text{pos}}}\left\| I^e-T^e \right\|^2_2 \\
\mathcal{L}_{\text{T,uniform}}\triangleq \log \mathop{\mathbb{E}}_{{\tiny{T_i, T_j}} \stackrel{\text{\tiny i.i.d.}}{\sim} p_{\text{data}}} e^{-2\left\| T_i^e-T_j^e \right\|^2_2} \\
\mathcal{L}_{\text{I,uniform}}\triangleq \log \mathop{\mathbb{E}}_{{\tiny{I_i, I_j}} \stackrel{\text{\tiny i.i.d.}}{\sim} p_{\text{data}}} e^{-2\left\| I_i^e-I_j^e \right\|^2_2}
\end{gathered}
\end{equation*}
where $(I,T) \sim p_{\text{pos}}$ refers to aligned
text-images pairs,
$(T_i,T_j) \sim p_{\text{data}}$ refers to
independent and identically distributed (IID) sampled text pairs,
$(I_i,I_j) \sim p_{\text{data}}$ refers to
IID sampled image pairs, and
$I^e$, $T^e$ respectively refer to
the encoded image and text vectors.

Recall that the trained models output L2 normalized
vectors residing on the unit ball.
Intuitively, we want vectors of aligned pairs of two modalities
to be well aligned in the representation space,
such that $\mathcal{L}_{\text{align}}$
is close to zero.
However, we want the space of
a single modality to be more uniform than
anisotropic \citep{ethayarajh-2019-contextual,wolfe-caliskan-2022-contrastive},
such that the overall representation space capacity is well used.
This results in a near minus infinite $\mathcal{L}_{\text{uniform}}$.

\begin{figure}
\centering
\subfloat[CLIP/CLIPs]{\includegraphics[width=0.242\textwidth]{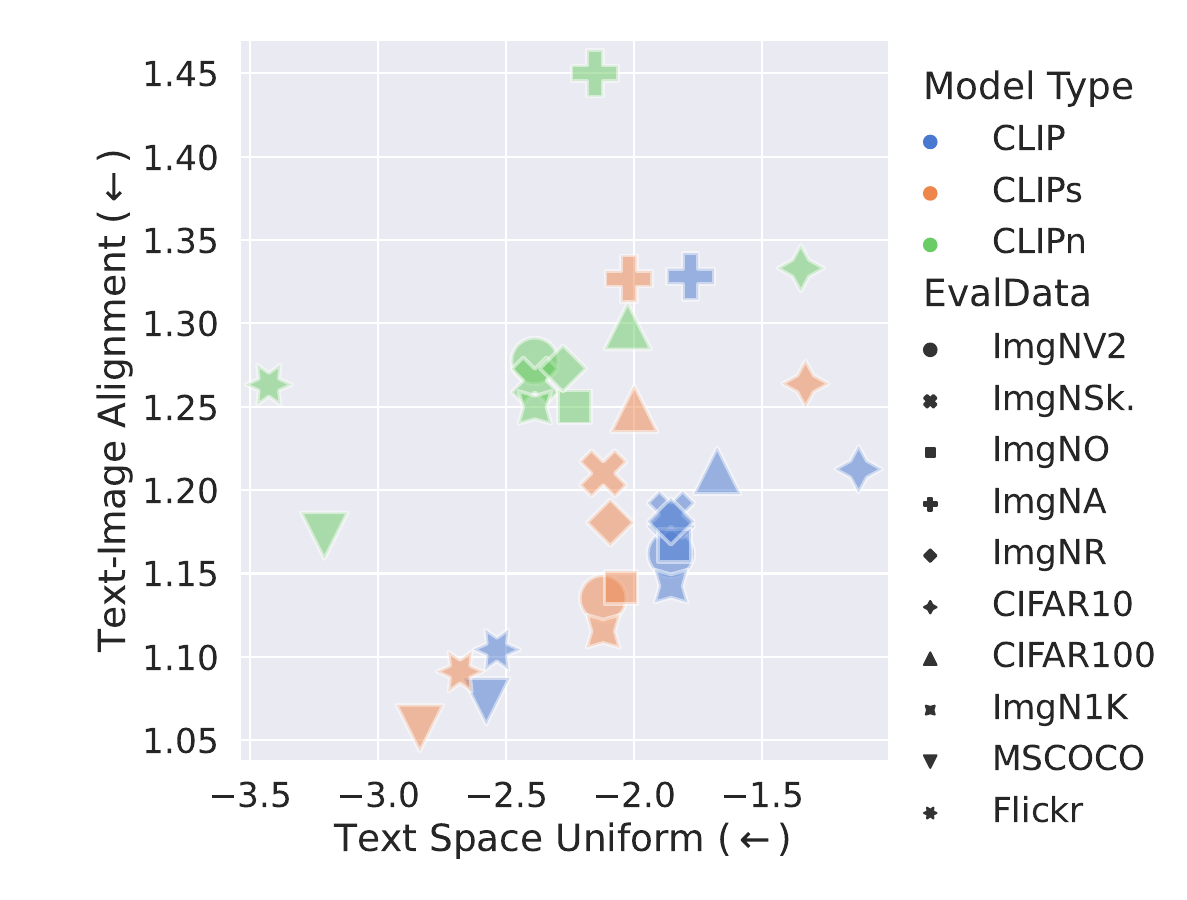}} \hspace{-.2cm}
\subfloat[CyCLIP/CyCLIPs]{\includegraphics[width=0.242\textwidth]{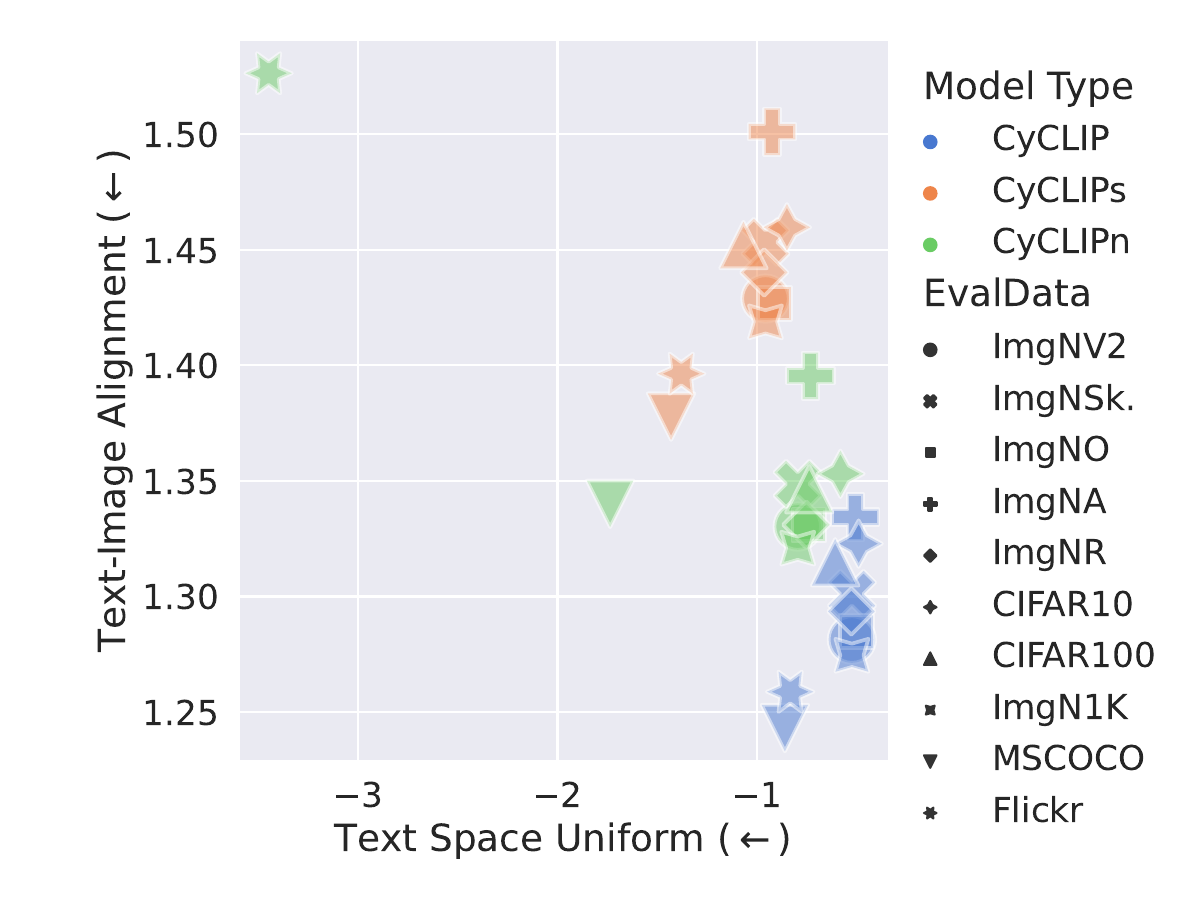}}\\
\subfloat[CLAP/CLAPs]{\includegraphics[width=0.24\textwidth]{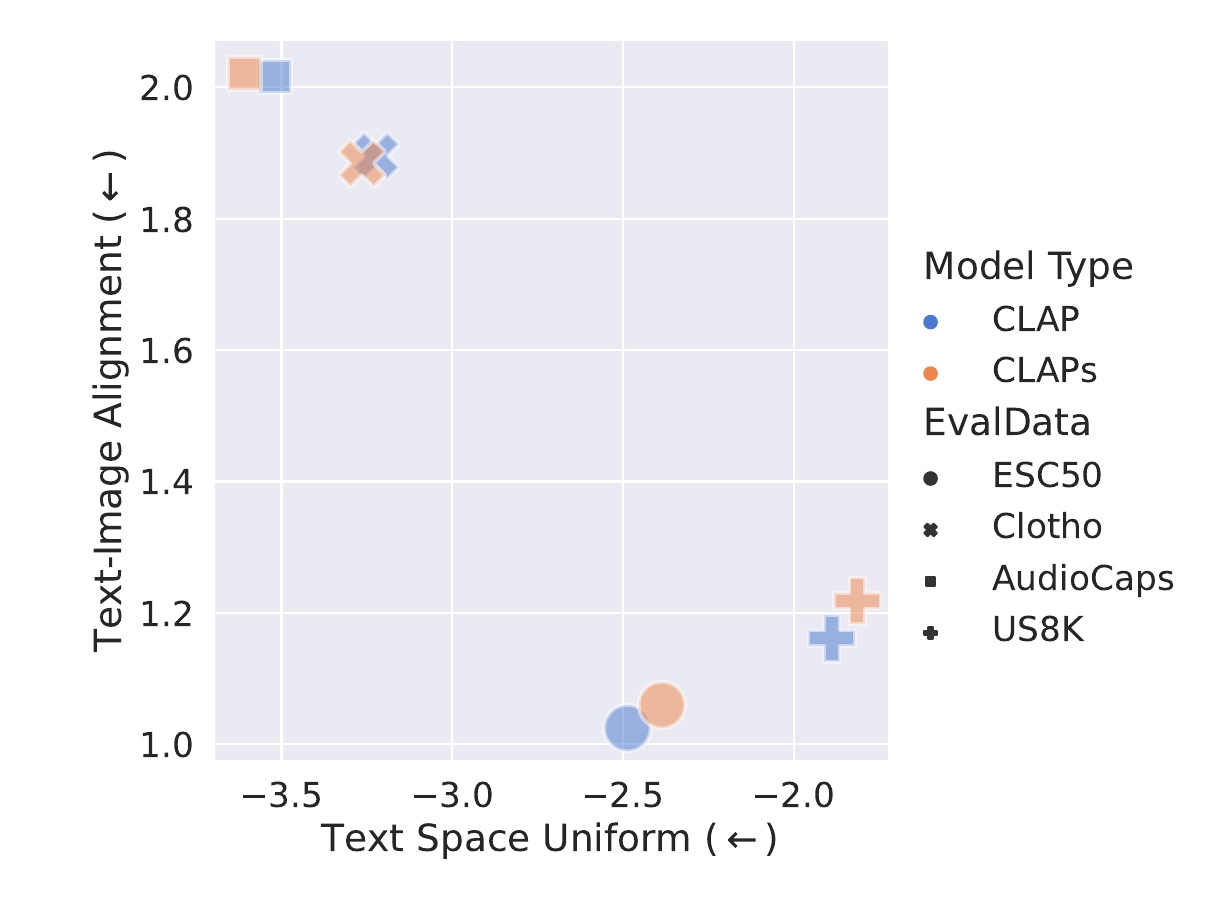}}\hspace{-.2cm}
\subfloat[CyCLAP/CyCLAPs]{\includegraphics[width=0.24\textwidth]{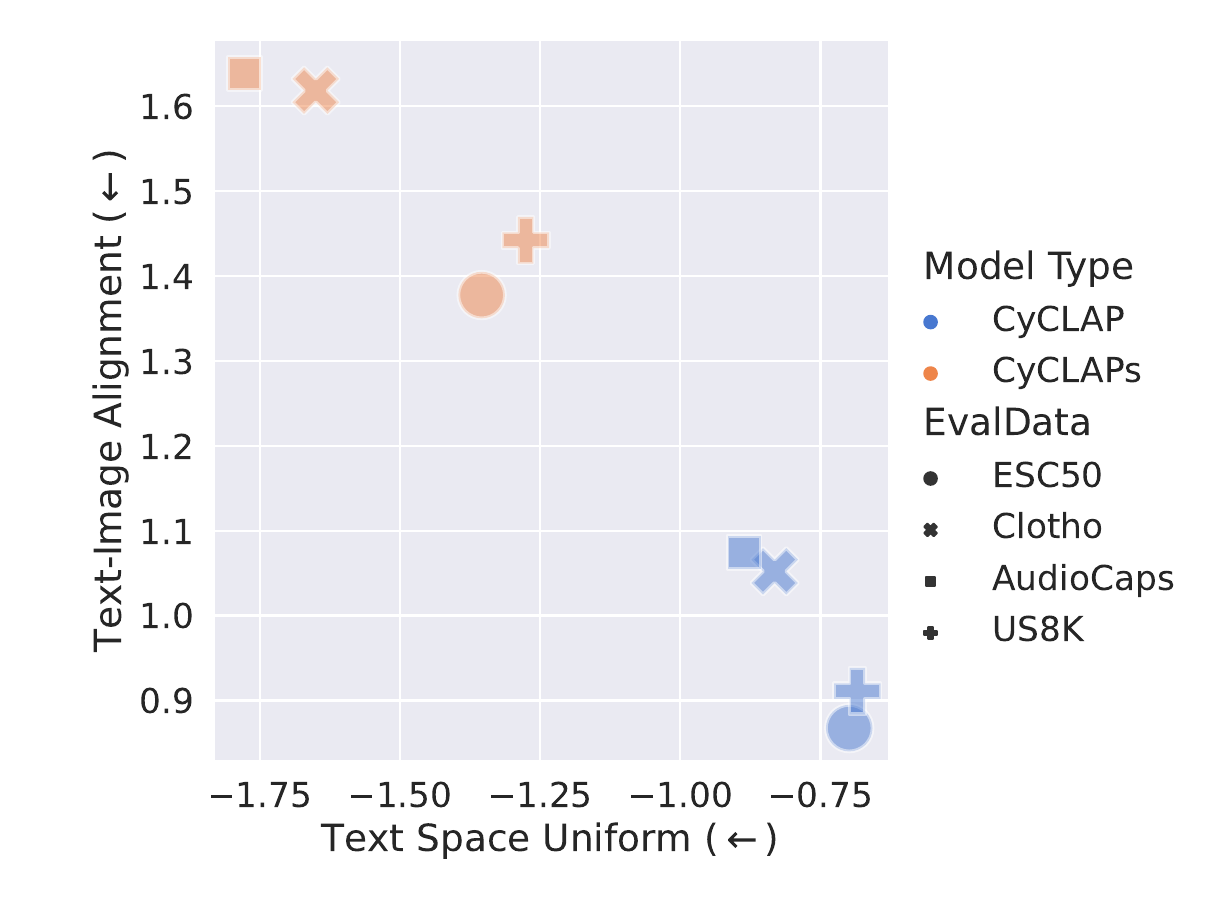}}
\caption{
Visualizing cross-modal alignment
w.r.t. text space uniformity of trained VL and AL models.
To visualize AL results, we use models pretrained
on AudioCaps due to it being larger than Clotho.
We observe that sentence embedding training
trades cross-modal alignment for text space uniformity.
}
\figlabel{aanduni}
\end{figure}

\figref{aanduni} illustrates the results.
We only show $\mathcal{L}_{\text{align}}$
w.r.t. $\mathcal{L}_{\text{T,uniform}}$ since
our main focus is the language encoder.
We see that unsupervised sentence
embedding training trades cross-modal
alignment for improving the
text space 
uniformity.
For VL pretraining,
supervised sentence embedding
training overly focuses on text space uniformity
while the VL space alignment deteriorates.
This observation is further evidenced when visualizing
with CyCLIP and CyCLIPn using the Flickr30K ($\varstar$) dataset
in which the captions largely overlap with the dataset for supervised sentence
embedding training (\secref{sec:expdataset}).

Another interesting observation is that the text encoder of
CyCLIP/CyCLAP outputs representation space less uniform than that of
CLIP/CLAP, as shown in \figref{aanduni} (x-axis).
\citet{liang2022mind} show that randomly initialized
encoders output vectors residing in different cones.
The in-modal cyclic loss $\mathcal{L_{\textnormal{I-cyclic}}}$
(\secref{sec:method}) stresses consistency between
cones; it is thus expected to be challenging to
learn a uniform space
while simultaneously  preserving consistency
between two spaces\footnote{
We find that improving text space uniformity
also benefits the image space for CyCLIP.
More discussions are presented in Appendix \secref{appendixsec:spaceconsistency}.
}.
Sentence embedding training provides extra training signals.

\textbf{LM quality}. We evaluate
the language encoder quality of
pretrained VL models.
Our motivation is two-fold.
First, as a sanity check, we want to
verify that incorporating sentence
embedding training in VL contrastive learning
still improves the language encoder's ability
of representing general sentences.
Second, the evaluation results help us
measure the compatibility and possible interferences
among the various training objectives \citep{pfeiffer2023modular}.

To this goal, we leverage the sentence embedding benchmark
SentEval \citep{conneau2018senteval}.
Default SentEval configurations
are used in all experiments.
Our intrinsic evaluation tasks
are the semantic textual similarity tasks:
STS12-STS16, STS-B, SICKR \citep{
marelli-etal-2014-sick,
cer-etal-2017-semeval,
agirre2012semeval,agirre2013sem,agirre2014semeval,
agirre2015semeval,agirre2016semeval}.
Extrinsic evaluation tasks are
movie review (MR; \citet{pang-lee-2005-seeing})
product review (CR; \citet{HuCR})
subjectivity status (SUBJ; \citet{pang-lee-2004-sentimental}),
opinion polarity (MPQA; \citet{wiebe2005annotating}),
sentiment analysis on SST2 \citep{socher2013recursive},
question-type classification (TREC; \citet{trec}),
and paraphrase detection (MRPC; \citet{dolan-etal-2004-unsupervised}).
\tabref{senteval} lists the results.
We observe that unsupervised sentence embedding training is beneficial
generally for CyCLIP/CLIP on both intrinsic and extrinsic tasks.
Supervised sentence embedding training results in more
significant improvements\footnote{
Improvements of
supervised sentence embedding training may due
to the observation that NLI datasets have similar domains
and language use
as SentEval tasks.
In Appendix \secref{appendxi:unsupwithnli}, we show that
the improvements are indeed from supervised training,
rather than domain similarity.
}, however, it negatively affects
CyCLIPn on the sensitive intrinsic tasks.

\begin{table}[t]
\centering
\scriptsize
\ra{1.1}
\begin{tabular}{@{}lrrr|rrr@{}}\toprule
           & CLIP  & CLIPn & CLIPs  & CyCLIP & CyCLIPn & CyCLIPs \\
\cmidrule{1-7}
STS12      & 46.14 & 54.25 & 50.31  & 37.84  & 45.60   & 40.42  \\
STS13      & 50.24 & 59.67 & 48.44  & 52.35  & 37.82   & 54.90  \\
STS14      & 48.70 & 59.26 & 51.73  & 46.58  & 40.55   & 49.46  \\
STS15      & 64.90 & 73.81 & 66.09  & 63.25  & 59.62   & 67.01  \\
STS16      & 51.94 & 63.08 & 55.62  & 50.96  & 46.80   & 52.87  \\
STS-B      & 61.54 & 68.36 & 65.04  & 60.30  & 54.88   & 60.72  \\
SICKR      & 64.70 & 73.09 & 65.82  & 64.78  & 62.62   & 64.34  \\
\emph{Avg} & 55.45 & \textbf{64.50} & 57.58  & 53.72  & 49.70   & \textbf{55.67}  \\\midrule
MR         & 61.07 & 63.66 & 61.11  & 59.51  & 62.78   & 60.65  \\
CR         & 67.63 & 71.07 & 68.03  & 67.02  & 73.67   & 66.12  \\
SUBJ       & 76.39 & 78.09 & 77.52  & 74.24  & 78.90   & 77.36  \\
MPQA       & 74.60 & 77.25 & 74.80  & 74.69  & 80.13   & 76.16  \\
SST2       & 61.67 & 66.89 & 63.65  & 61.50  & 68.42   & 64.25  \\
TREC       & 60.80 & 56.00 & 60.60  & 55.80  & 53.80   & 62.80  \\
MRPC       & 67.07 & 67.77 & 67.77  & 68.00  & 68.75   & 68.29  \\
\emph{Avg} & 67.03 & \textbf{68.68} & 67.64  & 65.82  & \textbf{69.49}   & 67.95 \\
\bottomrule
\end{tabular}
\caption{
Intrinsic (top) and extrinsic (bottom)
SentEval task performance of
the language encoder in VL
models.
}
\tablabel{senteval}
\end{table}
\normalsize

In this section, we have shown that (1) Sentence embedding training improves
text-space uniformity with decreased cross-modal alignment.  Better
balancing the two goals helps improve retrieval performance. (2)
Sentence embedding training improves language encoder quality on
SentEval, resonating the improved text-space uniformity.

\section{Conclusion}
We extensively evaluate
the effectiveness of sentence
embedding training for pretraining contrastive
vision-language
and audio-language models.
We show that it improves vision-language pretraining,
resulting in a better CyCLIP. However,
the benefits to audio-language pretraining are
noisy and less noticeable.
We conduct comprehensive analyses
and show that
sentence embedding training
increases text space uniformity,
but with a cost of
reduced cross-modal alignment.

\section{Limitations}
We acknowledge a few limitations that should be
considered.

We restrict our scope to cross-modal contrastive models involving
three specific modalities: language, image, and audio. While
contrastive learning has been successfully extended to other
modalities such as music, incorporating music poses additional
challenges, particularly regarding licensing and the heterogeneity of
music sources. Downloading music from the internet and mining reliable
music-language pairs are time-consuming tasks, which we did not
consider in detail in this study. Nevertheless, we conducted
initial experiments on the music
modality using MusicCaps \citep{agostinelli2023musiclm}.
The results are shown in \secref{appendxi:music}.

In our audio-language pretraining experiments, we explored both
pretraining from scratch and pretraining from publicly available
language and audio encoders. However, we believe that a more promising
direction would involve adapting the pretrained language encoder to
the audio domain by performing additional pretraining on audio
descriptions before engaging in cross-modal contrastive
learning. Nevertheless, we chose to follow the current methods in the
literature to ensure consistent evaluations and facilitate meaningful
comparisons.

\bibliography{anthology,custom}
\bibliographystyle{acl_natbib}

%\clearpage
\appendix
\section{Appendix}
\subsection{Datasets and hyperparameters}
\seclabel{appendixsec:dataandhyper}

For VL pretraining, our experiments largely follow those for CyCLIP
\citep{goel2022cyclip} and for CLIP \citep{radford2021learning}.  We
also directly reuse CyCLIP training hyperparameters but with a smaller
batch size, as listed in \tabref{vlhyper}.

For AL pretraining, our experiments largely follow
that of for LAION-CLAP \citep{laionclap2023}. To process the audio data,
we sample the wavefiles at a rate of 48kHz and then convert
them to FLAC format using \texttt{FFmpeg}\footnote{https://ffmpeg.org/}.
We then use a hop size of 480, window size of 1024, and 64 mel-bins
for computing Short-time Fourier transform (STFT) and
mel-spectrograms.  The audio encoder input thus
has a dimension of 1024 for time steps and 64 for frequency bins.
We list the hyperparameters in \tabref{alhyper}.

\begin{table}[t]
\begin{center}
\begin{tabular}{ll}
\toprule
Hyperparameter & Value \\
\midrule
Logit scale range & 0 to 4.6052 \\
Epochs & 64 \\
Batch size & 80 \\
Learning rate & 0.0005 \\
Optimizer & Adam \\
Scheduler & Cosine \\
Learning rate warmup steps & 10000 \\
Language encoder dropout & 0.1 \\
\bottomrule
\end{tabular}
\end{center}
\caption{Hyperparameters used for training VL models}
\tablabel{vlhyper}
\end{table}

\begin{table}[t]
\begin{center}
\begin{tabular}{ll}
\toprule
Hyperparameter & Value \\
\midrule
Logit scale range & 0 to 4.6052 \\
Epochs & 90 \\
Batch size & 80 \\
Learning rate & 0.00009 \\
Optimizer & AdamW \\
Scheduler & Cosine \\
Learning rate warmup steps & 9600 \\
Language encoder dropout & 0.1 \\
\bottomrule
\end{tabular}
\end{center}
\caption{Hyperparameters used for training AL models}
\tablabel{alhyper}
\end{table}

\subsection{Text and image space consistency}
\seclabel{appendixsec:spaceconsistency}
CyCLIP variants ensure cross-modal consistency, such that improving
the uniformity of the text space with sentence embedding training
also benefits image space uniformity
as shown in \figref{spaceconsistency}.
As expected, this observation does not hold for CLIP.

\begin{figure*}
\centering
\subfloat[CLIP: alignment vs. text space uniformity]{\includegraphics[width=0.5\textwidth]{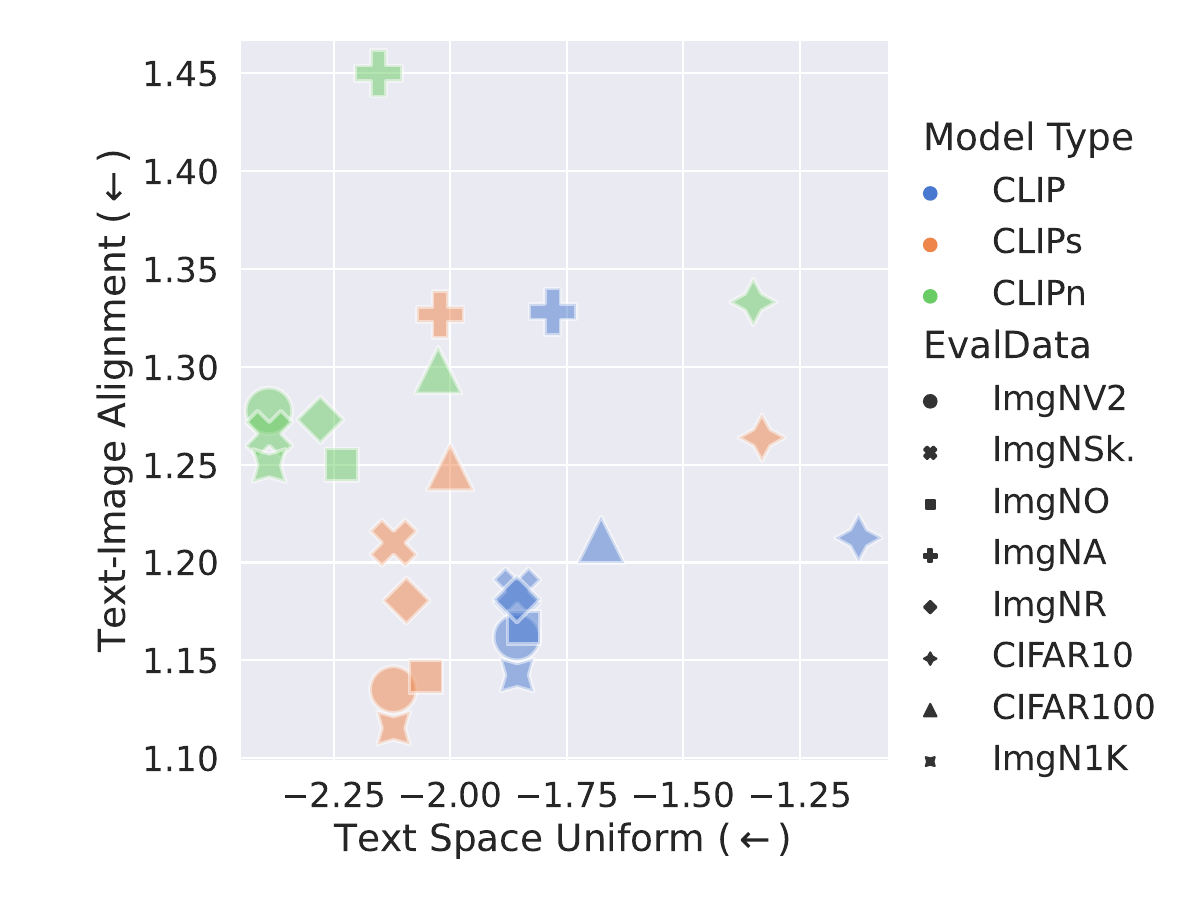}}
\subfloat[CLIP: alignment vs. image space uniformity]{\includegraphics[width=0.5\textwidth]{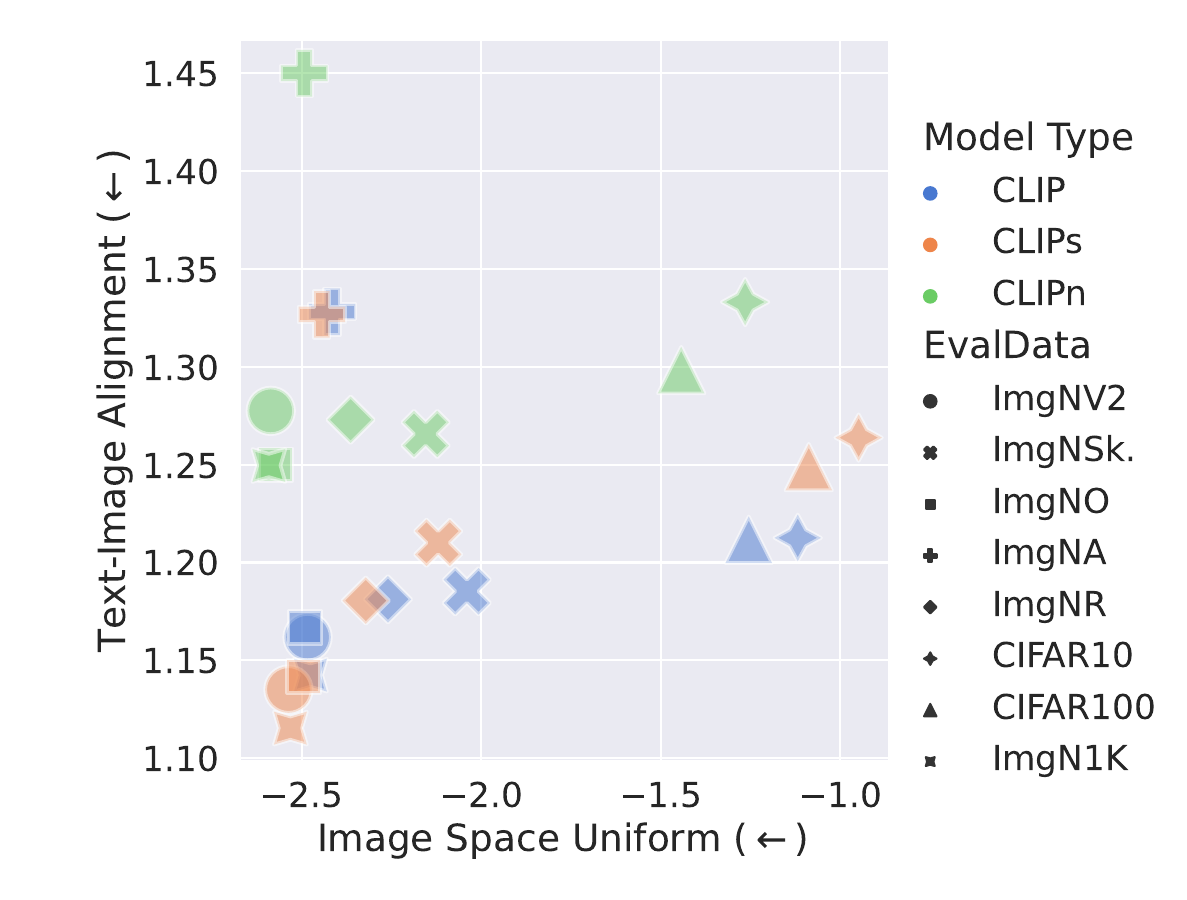}}\\
\subfloat[CyCLIP: alignment vs. text space uniformity]{\includegraphics[width=0.5\textwidth]{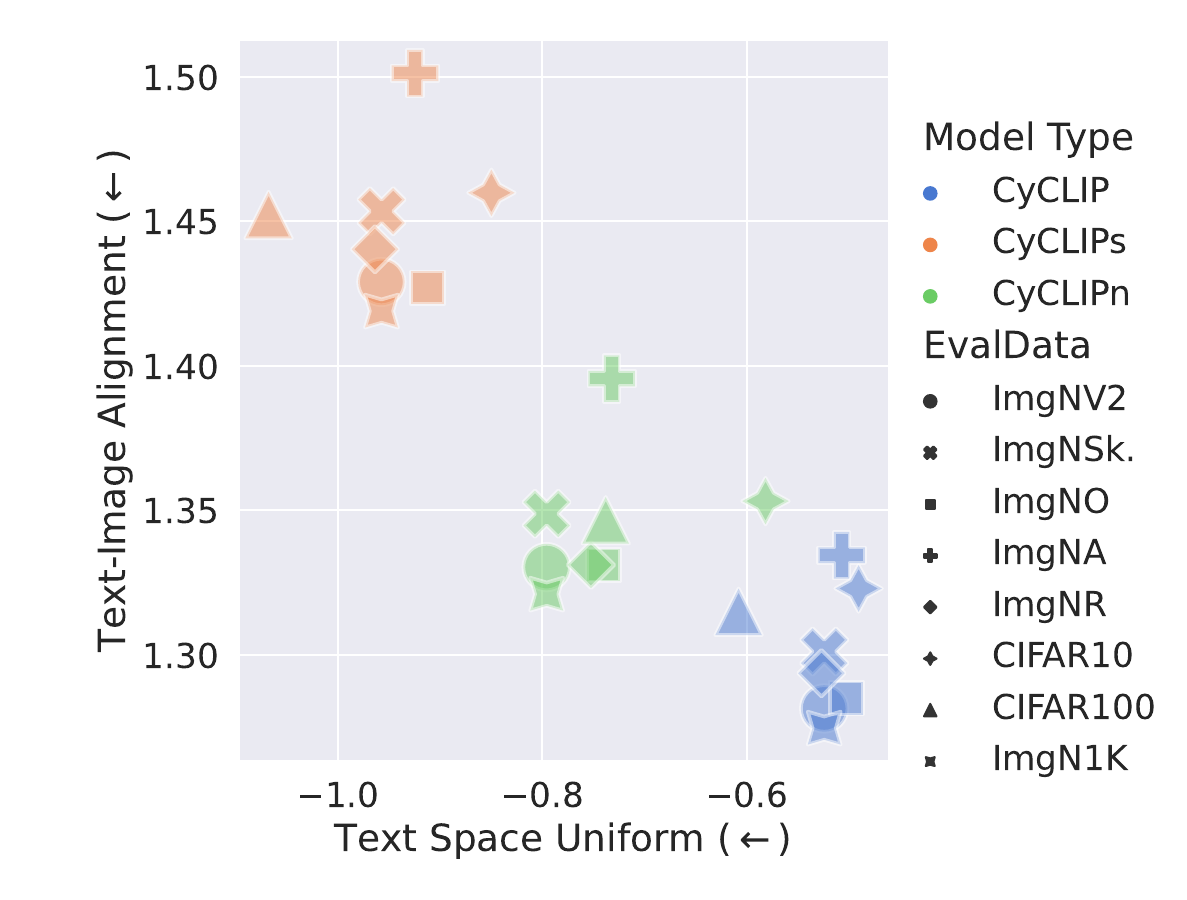}}
\subfloat[CyCLIP: alignment vs. image space uniformity]{\includegraphics[width=0.5\textwidth]{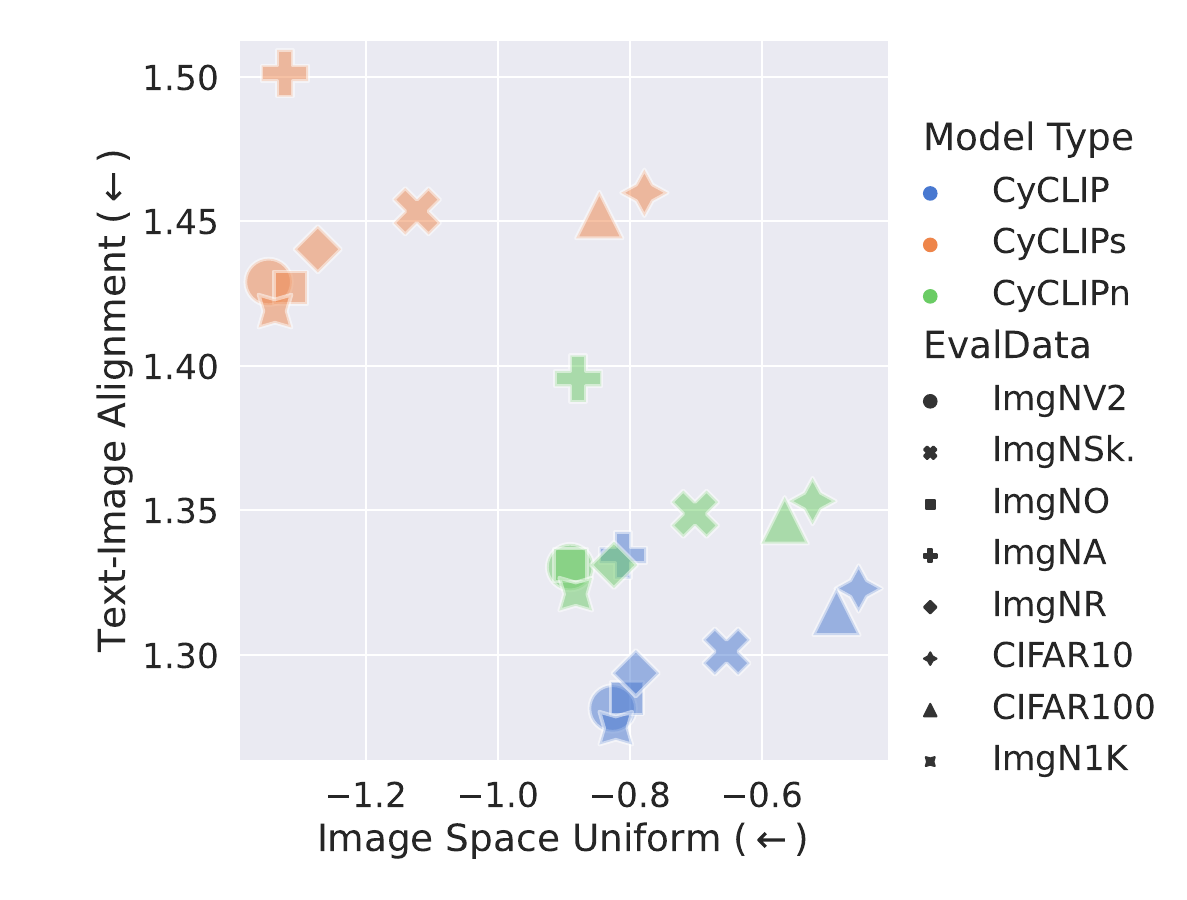}}
\caption{
Comparing the text and image space consistency between CLIP and CyCLIP variants.
Improving uniformity of the text space also benefits image space in CyCLIP.
}
\figlabel{spaceconsistency}
\end{figure*}

\subsection{Zero-shot audio classification with more prompts}
\seclabel{appendixsec:audiozeronewprompt}

To resemble zero-shot image classification in VL experiments, we write
more prompts for zero-shot audio classification, as listed in
\tabref{moreaudioprompts}.
\tabref{morepromptaudiozeroshotclassification} shows zero-shot audio
classification with these prompts.

It can be observed that (1) results in
\tabref{morepromptaudiozeroshotclassification} clearly outperform
\tabref{audiozeroshotclassification} results, illustrating the
importance of using many prompts; (2) sentence embedding training
improves the models trained on Clotho.

\begin{table}[t]
\ra{1.3}
\scriptsize
\centering
\begin{tabular}{@{}llrr|rr@{}}
\toprule
                           &       & CLAP  & CLAPs & CyCLAP & CyCLAPs \\ \midrule
\multirow{2}{*}{Clotho}    & ESC50 & 73.00 & \textbf{73.25} & 77.00  & \textbf{77.75}      \\
                           & US8K  & 69.64 & \textbf{70.27} & \textbf{71.82}  & 68.75      \\ \midrule
\multirow{2}{*}{AudioCaps} & ESC50 & \textbf{79.25} & 75.00 & \textbf{80.50}  & 78.50        \\
                           & US8K  & \textbf{72.00} & 66.49 & \textbf{69.46}  & 69.38     \\ \bottomrule
\end{tabular}
\caption{
Zero-shot audio classification (R@1 in \%)
on ESC50 and US8K. Prompts in \tabref{moreaudioprompts} are used.
}
\tablabel{morepromptaudiozeroshotclassification}
\end{table}

\begin{table}[t]
\centering
\begin{tabular}{@{}l@{}}
\toprule
    A sound of \texttt{label} \\
    a sound of \texttt{label}. \\
    A constant sound of \texttt{label}\\
    a constant sound of \texttt{label}\\
    A constant sound of \texttt{label}. \\
    a constant sound of \texttt{label}. \\
    A big sound of \texttt{label} \\
    a big sound of \texttt{label} \\
    A big sound of \texttt{label}. \\
    A small sound of \texttt{label} \\
    a small sound of \texttt{label} \\
    A small sound of \texttt{label}. \\
    A \texttt{label} is making a sound. \\
    a \texttt{label} is making a sound \\
    An \texttt{label} is making a sound. \\
    an \texttt{label} is making a sound \\
    A sound of \texttt{label} followed by a sound of \texttt{label} \\
    A sound of \texttt{label} followed by \texttt{label} \\
    A \texttt{label}@ \\
    A \texttt{label} \\
    An \texttt{label} \\
    an \texttt{label} \\
    \texttt{label}@ \\
    \texttt{label} \\
    \texttt{label} and \texttt{label} \\
    A \texttt{label} is running. \\
    A \texttt{label} is running \\
    A \texttt{label} is happening. \\
    A \texttt{label} is happening \\ \bottomrule

\end{tabular}
\caption{
We prepare additional prompts for zero-shot audio classification
resembling the VL prompts.
``\texttt{label}'' refers to the audio class label.
When a prompt contains
whitespace, we display it with ``@'' for better visualization.
}
\tablabel{moreaudioprompts}
\end{table}

\begin{table}[t]
\scriptsize
\begin{subtable}[b]{0.4\textwidth}
\hspace{-.15cm}
\ra{1.3}
\begin{tabular}{@{}lrrrp{.05cm}rrrc@{}}\toprule
     & \multicolumn{3}{c}{Text Retrieval} & \phantom{abc}& \multicolumn{3}{c}{Audio Retrieval}  \\
\cmidrule{2-4} \cmidrule{6-8}
         & R@1   & R@5   & R@10  && R@1   & R@5   & R@10  \\ \midrule
CLAP     & \textbf{2.30} & 7.85 & 13.88 && 2.28 & 7.94 & 13.47 \\
CLAPs    & 2.11 & \textbf{8.32} & \textbf{15.98} && \textbf{2.81} & \textbf{8.84} & \textbf{14.91} \\ \midrule
CyCLAP   & 2.97 & 9.09 & 14.07 && 2.07 & 7.67 & 13.11 \\
CyCLAPs  & \textbf{3.54} & \textbf{10.05} & \textbf{15.41} && \textbf{2.64} & \textbf{8.88} & \textbf{14.87} \\
%%\bottomrule
\end{tabular}
\end{subtable}
\vspace{1em}\quad\\
\begin{subtable}[b]{0.4\textwidth}
\hspace{-.15cm}
\ra{1.3}
\begin{tabular}{@{}lrrrp{.05cm}rrrcrrr@{}}%%\toprule
     & \multicolumn{3}{c}{Text Retrieval} & \phantom{abc}& \multicolumn{3}{c}{Audio Retrieval}  \\
\cmidrule{2-4} \cmidrule{6-8}
         & R@1   & R@5   & R@10  && R@1   & R@5   & R@10  \\ \midrule
CLAP     & 16.30 & 43.78 & \textbf{58.41} && \textbf{14.19} & \textbf{40.02} & \textbf{53.98} \\
CLAPs    & \textbf{17.76} & \textbf{44.10} & 57.99 && 13.81 & 38.60 & 52.02\\\midrule
CyCLAP   & \textbf{18.81} & 44.83 & \textbf{61.23} && \textbf{14.96} & 39.94 & 54.67\\
CyCLAPs  & 18.18 & \textbf{47.23} & 61.02 && 13.96 & \textbf{40.56} & \textbf{54.84}\\
\bottomrule
\end{tabular}
\end{subtable}
\caption{
Text and audio retrieval results (\%)
on Clotho (top) and AudioCaps (bottom)
when pretraining AL models from scratch.
}
\tablabel{scratchaudioretrievalres}
\end{table}
\normalsize

\subsection{Training audio-language models from scratch}
\seclabel{appendix:scratchaudio}
In \secref{subsec:res} we show that
sentence embedding training
brings more noticeable impacts
in learning VL models than
in AL models;
we conjecture that this is resulted from the fact that
AL pretraining
often leverages pretrained language
and audio encoders \citep{msclap,laionclap2023}.
As a result, we
conduct the experiments
of pretraining the AL model from scratch, i.e.,
the language and audio encoders re-initialized.
\tabref{scratchaudioretrievalres} lists the retrieval
results on Clotho and AudioCaps.

Compared with \tabref{audioretrievalres}, we observe a significant
performance drop since the encoders are pretrained from scratch.  The
results are still noisy. We consider that larger scale AL datasets are
necessary to highlight the effectiveness of learning consistent
representation spaces and sentence embedding training.

\subsection{Unsupervised sentence embedding training with NLI datasets}
\seclabel{appendxi:unsupwithnli}
When evaluting the language encoder
on SentEval tasks (\secref{alignmentuniform}),
it is possible that the improvements
brought by
supervised sentence embedding training
is due to the fact that NLI datasets have
similar domain and language use as the SentEval tasks.
We thus condcut a new type of training, where
we use sentences in the NLI datasets for unsupervised
sentence embedding training with SimCSE, in addition
to VL contrastive learning. We name this new training scheme
as CLIPe and CyCLIPe.

\begin{table}[t]
\scriptsize
\centering
\ra{1.3}
\begin{tabular}{@{}lrrrp{.15cm}rrrc@{}}\toprule
     & \multicolumn{3}{c}{Text Retrieval} & \phantom{abc}& \multicolumn{3}{c}{Image Retrieval}  \\
\cmidrule{2-4} \cmidrule{6-8}
         & R@1   & R@5   & R@10  && R@1   & R@5   & R@10  \\ \midrule
CLIP     & 15.70 & 37.22 & 49.06 && 12.48 & 31.10 & 42.23 \\
CLIPs    & \textbf{17.78} & \textbf{38.92} & \textbf{50.10} && \textbf{13.46} & \textbf{32.93} & \textbf{44.09} \\
CLIPn    & 15.74 & 35.66 & 47.38 && 13.12 & 31.46 & 42.55 \\
\textbf{CLIPe}    & 16.90 & 38.40 & 49.92 && 13.21 & 31.21 & 42.26 \\\midrule
CyCLIP   & 18.92 & 41.46 & 54.00 && 15.40 & 35.61 & 46.95 \\
CyCLIPs  & \textbf{21.30} & \textbf{44.34} & \textbf{56.54} && \textbf{16.69} & \textbf{37.75} & \textbf{49.24} \\
CyCLIPn  & 16.32 & 36.76 & 48.16 && 14.53 & 34.07 & 45.52 \\
\textbf{CLIPe}  & 16.22 & 37.52 & 49.22 && 14.05 & 32.56 & 43.15 \\
\bottomrule
\end{tabular}
\caption{
Zero-shot \textbf{VL retrieval results} (\%)
on MSCOCO (top) and Flickr30K (bottom).
}
\tablabel{retrievalreswitheunsupwithnli}
\end{table}
\normalsize

\tabref{sentevalwitheunsupwithnli} shows that the new training schemes,
CLIPe and CyCLIPe fall behind the supervised sentence embedding
training counterparts CLIPn and CyCLIPn on SentEval. This confirms that
the gains of supervised sentence embedding trainings
is from the NLI task supervision, e.g., premise and hypothesis relations,
instead of other factors such as domain.
For completeness, we also report CLIPe/CyCLIPe performance on VL
retrieval tasks in \tabref{retrievalreswitheunsupwithnli} and
zero-shot image classification in
\tabref{zeroshotclassificationunsupwithnli}.

\subsection{Preliminary experiments on the music modality}
\seclabel{appendxi:music}
We further conducted a new experiment with the music modality:
music-text retrieval on the MusicCaps dataset introduced by MusicLM
\citep{agostinelli2023musiclm}.  MusicCaps consists of 5521
music-caption pairs, of which 2858 pairs are for training and 2663 are
for validation. Each music clip is associated with hand-curated
English descriptions (including genre, mood, tempo, singer voices
etc.) from expert musicians. We use MusicCaps because it is
open-sourced and publicly available. Following tables show the
retrieval results (the same experiment configurations
as the audio modality are used; cf. \secref{appendixsec:dataandhyper}).

It can be observed from \tabref{musicmodalityres} that improving the text encoder with unsupervised
sentence embedding training also helps music-text retrieval in the
music modality, especially in the text retrieval scenario (CLAPs
generally outperforms CLAP; CyCLAPs generally outperforms
CyCLAP). These music modality results are consistent with our previous
findings on the image and audio modalities, and plan to explore more
in this direction in future work.

\begin{table}[t]
\scriptsize
\begin{subtable}[b]{0.4\textwidth}
\hspace{-.15cm}
\ra{1.3}
\begin{tabular}{@{}lrrrp{.05cm}rrrc@{}}\toprule
     & \multicolumn{3}{c}{Text Retrieval} & \phantom{abc}& \multicolumn{3}{c}{Music Retrieval}  \\
\cmidrule{2-4} \cmidrule{6-8}
         & R@1           & R@5            & R@10           && R@1           & R@5            & R@10  \\ \midrule
CLAP     & \textbf{6.05} & 18.42          & 28.33          && 5.46          & 18.45          & \textbf{28.50} \\
CLAPs    & 5.99          & \textbf{18.66} & \textbf{28.85} && \textbf{6.05} & \textbf{19.01} & 28.43 \\ \midrule
CyCLAP   & \textbf{6.37} & 18.98          & 29.45          && \textbf{6.69} & 18.87          & \textbf{29.45} \\
CyCLAPs  & 6.34          & \textbf{20.06} & \textbf{30.08} && 6.62          & \textbf{19.36} & 29.13 \\
\bottomrule
\end{tabular}
\end{subtable}
\vspace{1em}\quad\\
\caption{Text and music retrieval results (\%) on MusicCaps.}
\tablabel{musicmodalityres}
\end{table}
\normalsize

\begin{table*}[t]
\centering
\scriptsize
\ra{1.3}
\begin{tabular}{@{}lrrrr|rrrr@{}}\toprule
           & CLIP  & CLIPn           & \textbf{CLIPe}   & CLIPs  & CyCLIP & CyCLIPn    & \textbf{CyCLIPe}     & CyCLIPs \\
\cmidrule{1-9}
STS12      & 46.14 & 54.25           & 46.54   & 50.31  & 37.84  & 45.60         & 42.03  & 40.42  \\
STS13      & 50.24 & 59.67           & 46.29   & 48.44  & 52.35  & 37.82         & 35.56  & 54.90  \\
STS14      & 48.70 & 59.26           & 47.66   & 51.73  & 46.58  & 40.55         & 27.57  & 49.46  \\
STS15      & 64.90 & 73.81           & 65.48   & 66.09  & 63.25  & 59.62         & 46.65  & 67.01  \\
STS16      & 51.94 & 63.08           & 52.39   & 55.62  & 50.96  & 46.80         & 33.83  & 52.87  \\
STS-B      & 61.54 & 68.36           & 60.99   & 65.04  & 60.30  & 54.88         & 44.63  & 60.72  \\
SICKR      & 64.70 & 73.09           & 62.86   & 65.82  & 64.78  & 62.62         & 47.53  & 64.34  \\
\emph{Avg} & 55.45 & \textbf{64.50}  & 54.60   & 57.58  & 53.72  & 49.70         & 39.69  & \textbf{55.67}  \\\midrule
MR         & 61.07 & 63.66           & 59.91   & 61.11  & 59.51  & 62.78         & 59.60  & 60.65  \\
CR         & 67.63 & 71.07           & 68.74   & 68.03  & 67.02  & 73.67         & 64.61  & 66.12  \\
SUBJ       & 76.39 & 78.09           & 75.86   & 77.52  & 74.24  & 78.90         & 74.16  & 77.36  \\
MPQA       & 74.60 & 77.25           & 73.54   & 74.80  & 74.69  & 80.13         & 73.95  & 76.16  \\
SST2       & 61.67 & 66.89           & 60.19   & 63.65  & 61.50  & 68.42         & 60.46  & 64.25  \\
TREC       & 60.80 & 56.00           & 56.60   & 60.60  & 55.80  & 53.80         & 57.60  & 62.80  \\
MRPC       & 67.07 & 67.77           & 67.83   & 67.77  & 68.00  & 68.75         & 67.48  & 68.29  \\
\emph{Avg} & 67.03 & \textbf{68.68}  & 66.10   & 67.64  & 65.82  & \textbf{69.49}& 65.41  & 67.95 \\
\bottomrule
\end{tabular}
\caption{Evaluating the language encoder of different VL
models with intrinsic (top) and extrinsic (bottom) SentEval tasks.
}
\tablabel{sentevalwitheunsupwithnli}
\end{table*}
\normalsize

\begin{table*}[t]
\centering
\tiny
\ra{1.3}
\begin{tabular}{@{}lrrrr|rrrr@{}}\toprule
                & CLIP  & CLIPn          & \textbf{CLIPe} & CLIPs          & CyCLIP & CyCLIPn & \textbf{CLIPe} & CyCLIPs \\
\cmidrule{1-9}
CIFAR10         & 28.31 & \textbf{44.06} & 33.97 & 36.80          & 38.67  & 41.16   & \textbf{50.48}   & 44.97 \\
CIFAR100        & 13.23 & \textbf{17.93} & 12.30 & 10.72          & 17.44  & 19.82   & 21.76   & \textbf{22.05} \\
ImageNet1K      & 14.94 & 15.97          & 15.74 & \textbf{16.01} & 20.99  & 18.13   & 20.07   & \textbf{22.13} \\\midrule
ImageNetV2      & 12.85 & 13.41          & 13.51 & \textbf{14.09} & 17.77  & 15.65   & 17.34   & \textbf{18.68} \\
ImageNet-Sk.    & 7.72  & 7.75           & 6.36  & \textbf{8.14}  & 11.67  & 9.93    & 11.54   & \textbf{12.85} \\
ImageNet-O      & 20.75 & \textbf{21.95} & 20.45 & 21.30          & 27.05  & 24.45   & 27.20   & \textbf{29.55} \\
ImageNet-A      & 3.59  & 3.41           & 3.59  & \textbf{3.95}  & 5.03   & 4.45    & 4.93   & \textbf{5.19} \\
ImageNet-R      & 18.39 & \textbf{18.51} & 18.25 & 18.24          & 24.37  & 23.07   & 24.36   & \textbf{26.72} \\
\bottomrule
\end{tabular}
\caption{
Zero-shot image classification (R@1 in \%) on
standard datasets (top) and
datasets with distribution shift
or adversarial examples (bottom).
}
\tablabel{zeroshotclassificationunsupwithnli}
\end{table*}
\normalsize

\end{document}